\documentclass[conference]{IEEEtran}
\IEEEoverridecommandlockouts
\usepackage{amsmath,amssymb,amsfonts}
\usepackage{algorithmic}
\usepackage{graphicx}
\usepackage{textcomp}
\usepackage{xcolor}
\usepackage{makecell}
\usepackage[style=ieee, citestyle=numeric-comp, backend=biber]{biblatex}
\AtBeginBibliography{\small}
\def\BibTeX{{\rm B\kern-.05em{\sc i\kern-.025em b}\kern-.08em
    T\kern-.1667em\lower.7ex\hbox{E}\kern-.125emX}}
\usepackage{authblk}
\usepackage{blindtext}
    
\begin{document}

\title{Face Photo-Sketch Recognition Using Bidirectional Collaborative Synthesis Network
\thanks{This work was supported by the National Research Foundation of Korea(NRF) grant
funded by the Korea government(MSIT) (No. 2020R1F1A1048438).
This research was results of a study on the "HPC Support" Project, supported by the ‘Ministry of Science and ICT’ and NIPA.}
}

\author[ ]{Seho Bae}
\author[ ]{Nizam Ud Din}
\author[ ]{Hyunkyu Park}
\author[ ]{Juneho Yi}
\affil[ ]{Department of Electrical and Computer Engineering}
\affil[ ]{Sungkyunkwan University}
\affil[ ]{Suwon, Republic of Korea}
\affil[ ]{\textit {\{bseho, nizam, mjss016, jhyi\}@skku.edu}}

\maketitle

\begin{abstract}
This research features a deep-learning based framework to address the problem of matching a given face sketch image against a face photo database. The problem of photo-sketch matching is challenging because 1) there is large modality gap between photo and sketch, and 2) the number of paired training samples is insufficient to train deep learning based networks. 
To circumvent the problem of large modality gap, our approach is to use an intermediate latent space between the two modalities. We effectively align the distributions of the two modalities in this latent space by employing a bidirectional (photo $\rightarrow$ sketch and sketch $\rightarrow$ photo) collaborative synthesis network. A StyleGAN-like architecture is utilized to make the intermediate latent space be equipped with rich representation power. 
To resolve the problem of insufficient training samples, we introduce a three-step training scheme. Extensive evaluation on public composite face sketch database confirms superior performance of our method compared to existing state-of-the-art methods. The proposed methodology can be employed in matching other modality pairs.
\end{abstract}


\begin{figure}
\begin{center}
\includegraphics[width=0.45\textwidth]{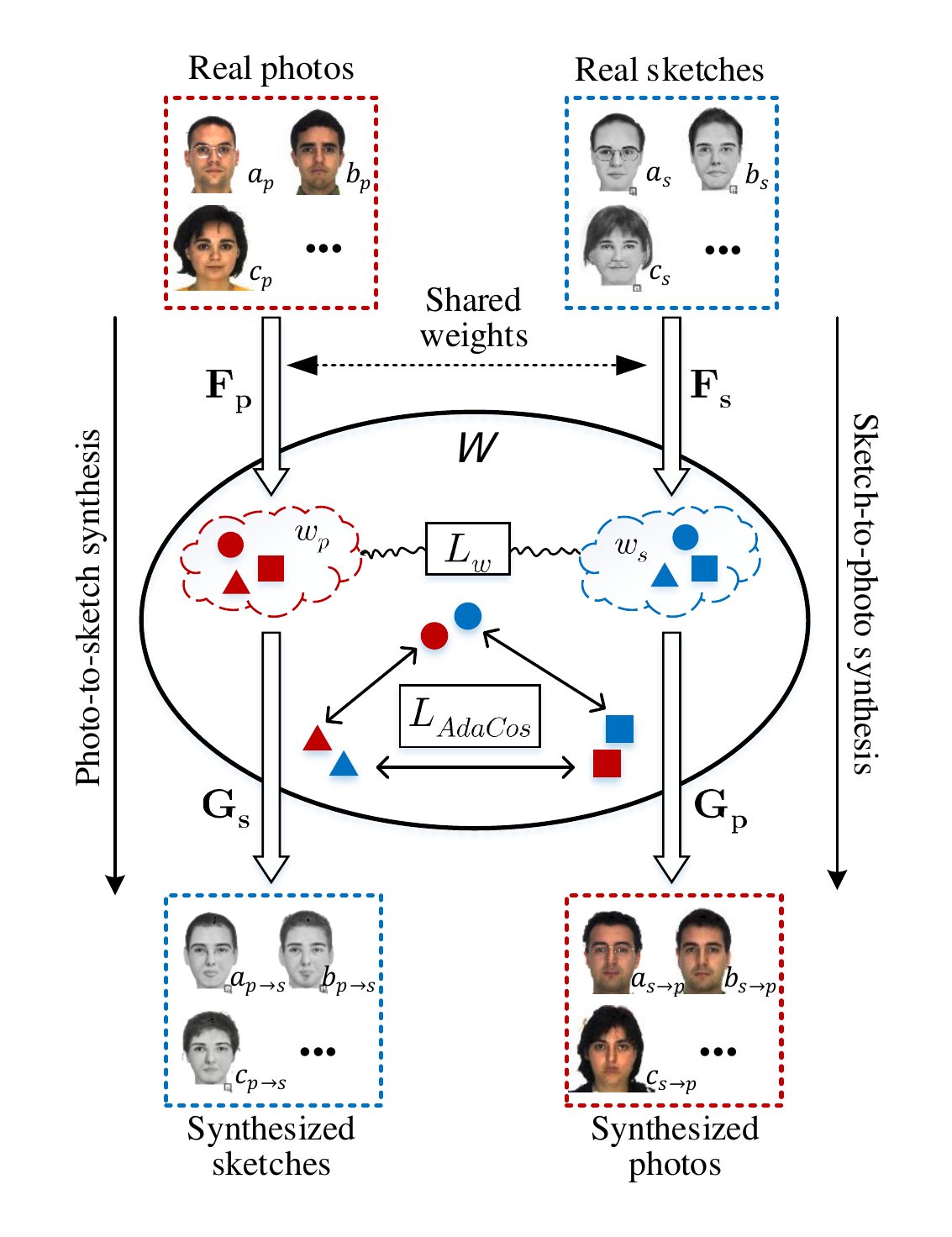}
\end{center}
\vspace{-1.5 em}
   \caption{Our proposed framework takes advantage of a bidirectional photo/sketch synthesis network to set up an intermediate latent space as an effective homogeneous space for face photo-sketch recognition. We employ a StyleGAN-like architecture to make the intermediate latent space be equipped with rich representational power. The mapping networks, $\mathbf F_{p}$ and $\mathbf F_{s}$, learn to encode photo and sketch images into their respective intermediate latent codes, $\mathit w_{p}$ and $\mathit w_{s}$. We learn AdaCos~\cite{zhang2019adacos} to enforce the separability of latent codes of different identity in the angular space  for the photo-sketch recognition task.
   }
\label{fig:framework}
\vspace{-1em}
\end{figure}

\section{Introduction}
\label{Section1}

The goal of this work is to find the best matching photos for a given sketch in a face database, especially for software generated composite sketches.
An important application of such systems is to assist law-enforcement agencies.
During criminal investigation, in many cases, facial photo of a suspect is not available. Instead, a hand-drawn forensic sketch or software generated composite sketch based on the description provided by an eye-witness or victim is the only clue to identify suspect. 
Therefore, an automatic method which retrieves the best matching photos from face database for a given sketch is necessary to quickly and accurately identify a suspect.

Successful photo-sketch matching depends on the solution to how to effectively deal with large modality gap between photos and sketches. Moreover, insufficiency of sketch samples for training makes photo-sketch recognition an extremely challenging task.

As to classical photo-sketch recognition, generative approaches~\cite{wang2008face,liu2005nonlinear,ouyang2016forgetmenot} bring both modalities into a single modality by transforming one of the modalities to the other (either photo to sketch or vice versa) before matching. The main drawback of these methods is their dependency on the quality of the synthetic output, which most of the time suffers due to large modality gap between the two modalities.
On the other hand, discriminative approaches attempt to extract modality-invariant features, or learn a common subspace where both photo and sketch modalities are aligned \cite{sharma2011bypassing,choi2012data,zhang2011coupled,klare2010matching,ouyang2014cross,mittal2014recognizing,han2012matching,bhatt2012memetically,peng2016graphical}. Although these methods formulate photo-sketch recognition through modality invariant features or a common subspace, their performances are not satisfactory because 1) the distributions of the two modalities are not well aligned in the common feature space and 2) their feature vectors or common spaces fail to provide rich representation capacity.
Recent deep-learning based face photo-sketch recognition methods~\cite{kazemi2018attribute, iranmanesh2018deep, liu2020coupled, liu109iterative, peng2019dlface, xu2020matching, zhang2011coupled, mittal2014recognizing, spnet} perform well compared to classical approaches. However, utilizing deep learning techniques for face photo-sketch recognition is very challenging because of insufficient training data.

Recently, Col-cGAN ~\cite{zhu2019gan} proposed a bidirectional face photo-sketch synthesis network. They generate synthetic outputs by using a middle latent domain between photo and sketch modalities. 
However, their middle latent domain does not provide enough representational power of both modalities.
On the other hand, StyleGAN~\cite{karras2019style} produces extremely realistic images by proposing a novel generator architecture. Instead of feeding the input latent code $\mathit{z}$ directly into the generator, the StyleGAN network first transforms it into an intermediate latent space, $\mathit{W}$, via a mapping network. This disentangled intermediate latent space, $\mathit{W}$, offers the StyleGAN generator more control and representational capabilities. Noting the strong representation power of the latent code space of StyleGAN, we opt to use a StyleGAN-like bidirectional architecture for setting up an intermediate latent space for our photo-sketch recognition problem.

In this paper, we propose a novel method that exploits an intermediate latent space, $\mathit{W}$, between the photo and sketch modalities as shown in Figure \ref{fig:framework}. We employ a bidirectional collaborative synthesis network of the two modalities to set up the intermediate latent space where the distributions of the two modalities are effectively aligned. Also, the StyleGAN-like architecture we utilize enables the intermediate latent space to have strong representational power to successfully match the two modalities. 

In Figure \ref{fig:framework}, the mapping networks, $\mathbf F_{p}$ and $\mathbf F_{s}$, learn the intermediate latent codes $\mathit w_{p}, w_{s} \in W$. To form a homogeneous intermediate space, $\mathit{W}$, we constrain the intermediate features more symmetrical, using $\mathit \ell_{1}$ distance between the intermediate latent codes of photo and sketch 
The intermediate latent space also makes use of feedback from the style generators that translate photo-to-sketch/sketch-to-photo. Hereby enabling the intermediate latent space to have rich representational capacity for both photo and sketch.
Once this intermediate latent space is successfully set up, we can then directly take advantage of any state-of-the-art face recognition methods. 
In our case, we employ AdaCos loss~\cite{zhang2019adacos}.

Moreover, we use a three-step training scheme to resolve the problem of very limited number of training sketch samples.
In the first step, we only learn image-to-image translation without AdaCos on paired photo-sketch samples.
This serves the purpose of learning an initial intermediate latent space.
Then, in the second step, we pre-train the photo mapping network, $\mathbf F_{p}$, only with AdaCos, using a publicly available large photo dataset.
This helps our model overcoming the problem of insufficient sketch samples to train our deep network robustly for the target task.
Lastly, we fine tune the full network on a target photo/sketch dataset. More details of the model training are discussed in section \ref{sec:training}.

The main contributions of our work are summarized as follows.

\begin{itemize}
     \item We propose a novel method for photo-sketch matching that exploits an intermediate latent space between the photo and sketch modalities: 
      \begin{itemize}
     \item The intermediate latent space is built through a bidirectional collaborative synthesis network.
     \item This latent space has rich representational power for photo/sketch recognition due to a StyleGAN-like architecture.
   \end{itemize}
    \item A three-step training scheme helps overcoming the problem of insufficient sketch training samples. 
    \item Extensive evaluation on challenging publicly available composite face sketch databases shows superior performance of our method compared with state-of-art methods.
\end{itemize}

The rest of this paper is organized as follows. Section \ref{Section2} describes related works. In section \ref{Section3}, we depict details of our method. Experimental results are presented in section \ref{Section4}.


\section{Related work}
\label{Section2}

The face photo-sketch recognition problem has been extensively studied in recent years. Researchers have studied sketch recognition for various face sketch categories such as hand-drawn viewed sketch, hand-drawn semi-forensic sketch, hand-drawn forensic sketch, and software-generated composite sketch. Compared to hand-drawn viewed sketches, other sketch categories have much larger modality gap due to the errors that come from forgetting (semi-forensic/forensic), understanding of description (forensic), or limitation of components in software (composite). Recent researches focus on more challenging composite and forensic sketches.

Traditional sketch recognition methods can be divided into two categories: generative and discriminative approaches. 

\begin{figure*}
\begin{center}
\includegraphics[width=1\textwidth]{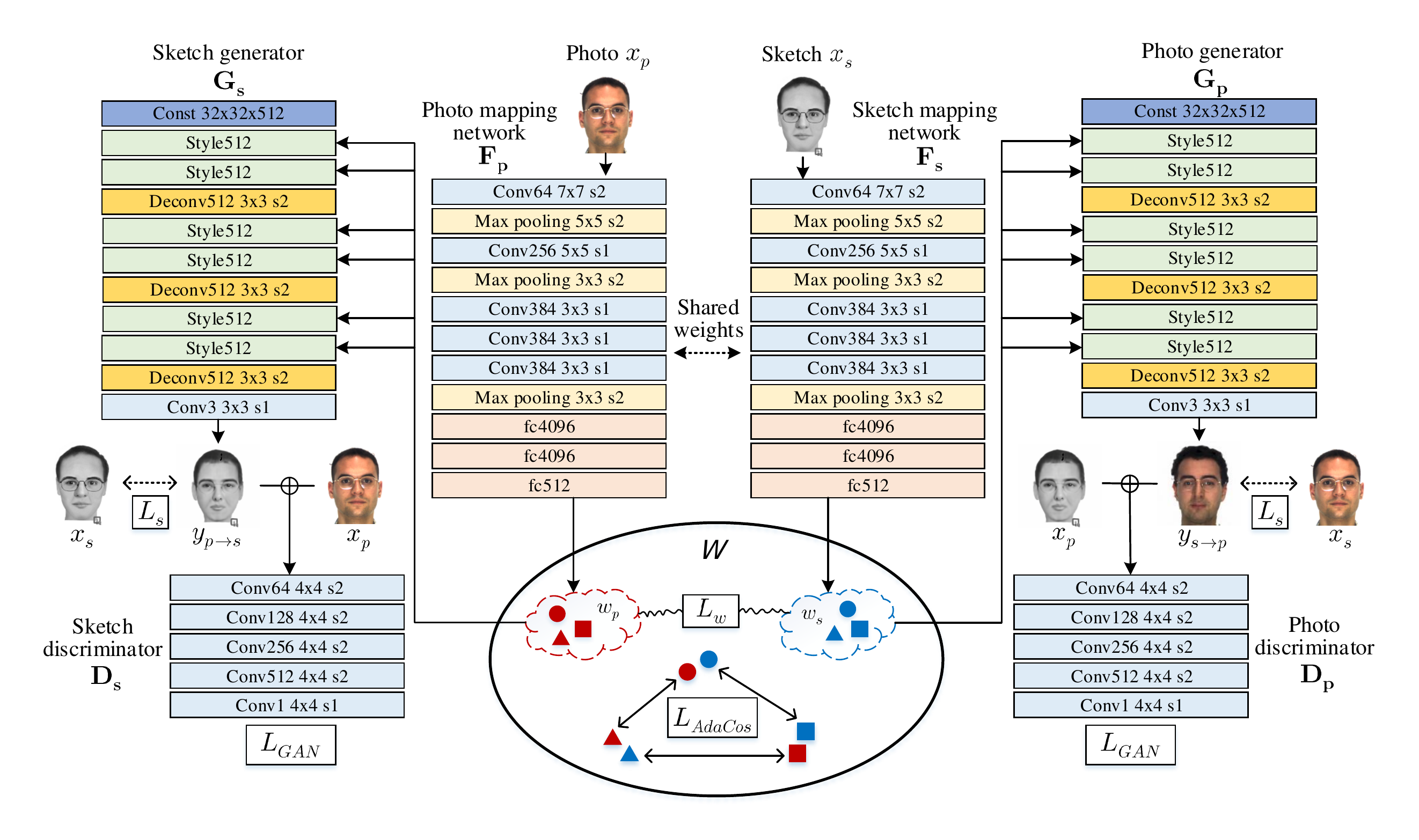}
\end{center}
\vspace{-1.5em}
   \caption{The overall architecture of the proposed network. Mapping networks, $\mathbf F_{p}$ and $\mathbf F_{s}$, map photo and sketch images to intermediate latent codes $\mathit w_{p}$ and $\mathit w_{s}$. These latent codes are then fed into the two opposite style generators $\mathbf G_{s}$ and $\mathbf G_{p}$. $\mathbf G_{s}$ generates sketch from photo, $\mathit y_{p\rightarrow s}$, while $\mathbf G_{s}$ generates photo from sketch, $\mathit y_{s\rightarrow p}$.
   The collaborative loss $\mathit L_{w}$, which is $\mathit \ell_{1}$ distance between $\mathit w_{p}$ and $\mathit w_{s}$ of same identity, constrains the intermediate modality features more symmetrical. Through this strategy, we learn an intermediate latent space, $\mathit{W}$, that retain the common and representational information of photo and sketch. We apply AdaCos loss, $\mathit L_{AdaCos}$, to the intermediate latent space, $\mathit{W}$, directly to perform photo-sketch recognition by comparing the cosine distance between intermediate latent features, $\mathit w_{p}$ and $\mathit w_{s}$.}
\label{fig:network}
\vspace{-1em}
\end{figure*}

Generative methods convert images from one modality into the other modality, usually from sketch to photo, before matching. Then, a simple homogeneous face recognition method can be used for matching. Various techniques have been utilized for synthesis such as Markov random field model~\cite{wang2008face}, local linear embeding (LLE)~\cite{liu2005nonlinear}, and multi-task gaussian process regression~\cite{ouyang2016forgetmenot}.
However, recognition performance of these methods heavily depends on the quality of the synthetic images, which most of the time suffers due to the large modality gap between the two modalities.

Discriminative methods attempt to learn a common subspace or extract particular features in order to reduce the intra-class difference caused by the modality gap while preserving inter-class separability. 
Representative methods in this category include partial linear square(PLS)~\cite{sharma2011bypassing,choi2012data}, coupled information-theoretic projection (CITP)~\cite{zhang2011coupled}, local feature-based discriminant analysis (LFDA)~\cite{klare2010matching}, canonical correlation analysis (CCA)~\cite{ouyang2014cross}, and self similarity descriptor (SSD) dictionary~\cite{mittal2014recognizing}. Han \textit{et al.}~\cite{han2012matching} proposed a component-based representation approach to measure the similarity between a composite sketch and photo. Multi-scale circular Weber's local descriptor (MCWLD) is utilized in Bhatt \textit{et al.}~\cite{bhatt2012memetically}  to solve semi-forensic and forensic sketch recognition problem. In graphical representation based heterogeneous face recognition (G-HFR)~\cite{peng2016graphical}, the authors graphically represented heterogeneous image patches by employing Markov networks, and designed a similarity metric for matching. 
These methods fail when the learned feature/common subspace could not have enough representational capacity for both photo and sketch modalities. 
In contrast, our method projects photo and sketch on homogeneous intermediate space where the distribution of the two modalities better aligned with rich representational power.

Over the past few years, deep learning based algorithms have been developed for face photo-sketch recognition~\cite{kazemi2018attribute, iranmanesh2018deep, liu2020coupled, liu109iterative, peng2019dlface, xu2020matching, zhang2011coupled, mittal2014recognizing}. Kazemi \textit{et al.}~\cite{kazemi2018attribute} and Iranmanesh \textit{et al.}~\cite{iranmanesh2018deep} proposed attribute-guided approaches by introducing attribute-centered loss function and joint loss function of identity and facial attribute classification, respectively. Liu \textit{et al.} designed coupled attribute guided triplet loss (CAGTL) to train an end-to-end network that can effectively eliminates defects of incorrectly estimated attributes~\cite{liu2020coupled}.  Iterative local re-ranking with attribute guided synthesis based on GAN is introduced in~\cite{liu109iterative}. Peng \textit{et al.} proposed DLFace~\cite{peng2019dlface} which is a local descriptor approach based on deep metric learning while in ~\cite{xu2020matching}, a hybrid feature model was employed by fusing traditional HOG feature with deep feature. The largest obstacle to utilizing deep learning techniques for face photo-sketch recognition is scarcity of sketch data. Even the largest public viewed sketch database~\cite{zhang2011coupled} has only 1,194 pairs of sketch and photo, and the composite sketch database~\cite{mittal2014recognizing} has photos and sketches of 123 identities. To overcome this problem, most approaches employ relatively shallow network, data augmentation, or pre-training on a large-scale face photo database.

Recently, cosine-based softmax losses~\cite{liu2017sphereface,wang2018cosface,deng2019arcface,zhang2019adacos} have achieved great success in face photo recognition. SphereFace~\cite{liu2017sphereface} penalises the angles between the deep features and their corresponding weights in a multiplicative way. Follow-up studies improved the performance by changing the penalising measure to additive margin in cosine~\cite{wang2018cosface} and angle~\cite{deng2019arcface}. AdaCos~\cite{zhang2019adacos} outperforms previous cosine-based softmax losses by leveraging an adaptive scale parameter to automatically strengthen the supervision during training. However, direct application of these methods to photo-sketch recognition is not satisfactory because  they have not properly dealt with the modality gap.

\begin{figure*}
\begin{center}
\includegraphics[width=1\textwidth]{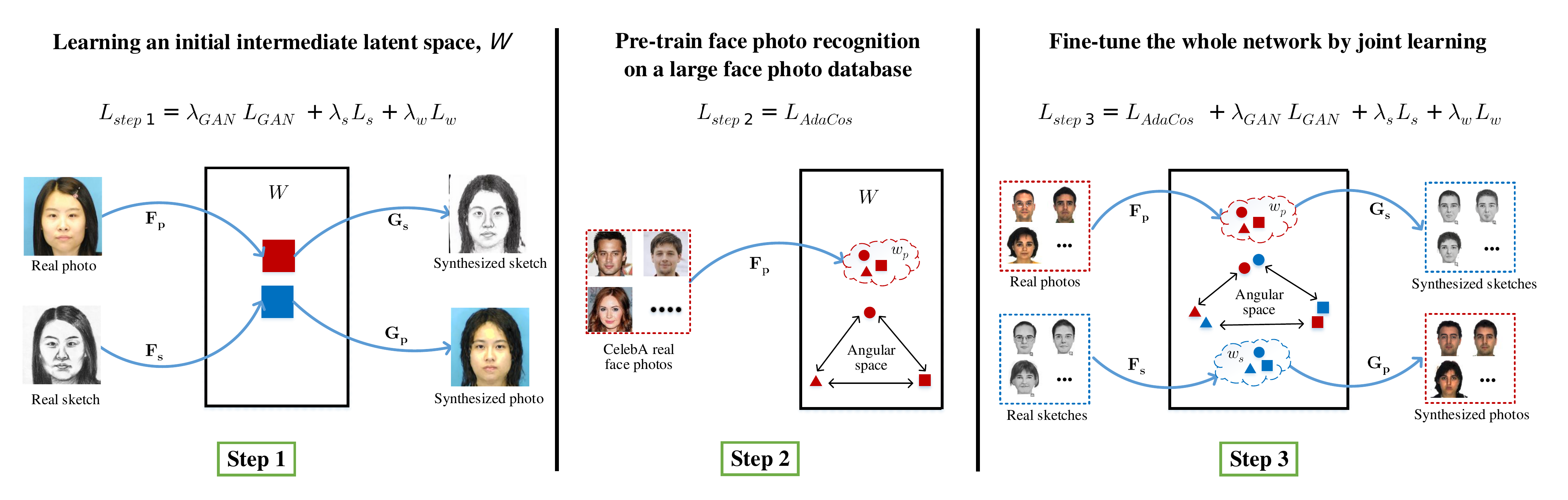}
\end{center}
   \caption{A three-step training scheme to overcome the problem of insufficient amount of paired photo-sketch training samples. We employ three-step training. 
   Step 1: Pre-train the bidirectional photo/sketch synthesis network to learn an initial intermediate latent space, $\mathit{W}$, between photo and sketch. Step 2: Pre-train the photo mapping network, $\mathbf F_{p}$, on a large face photo database.
   Step 3: Fine-tune the whole network on the target photo/sketch database.}
\label{fig:training}
\vspace{-1em}
\end{figure*}
\section{Proposed method}
\label{Section3}

Our proposed framework takes advantage of a bidirectional photo/sketch synthesis network to set up an intermediate latent space as an effective homogeneous space for face photo-sketch recognition. Mutual interaction of the two opposite synthesis mappings occurs in the bidirectional collaborative synthesis network.  The complete structure of our network is illustrated in Figure 2. Our network consists of mapping networks $\mathbf F_{p}$ and $\mathbf F_{s}$, style generators $\mathbf G_{p}$ and $\mathbf G_{s}$, and discriminators $\mathbf D_{p}$ and $\mathbf D_{s}$. $\mathbf F_{p}$ and $\mathbf F_{s}$ share their weights.

The mapping networks, $\mathbf F_{p}$ and $\mathbf F_{s}$, learn to encode photo and sketch images into their respective intermediate latent codes, $\mathit w_{p}$ and $\mathit w_{s}$. Then, $\mathit w_{p}$ and $\mathit w_{s}$ are fed into the two opposite style generators $\mathbf G_{s}$ and $\mathbf G_{p}$ to map photo-to-sketch and sketch-to-photo, respectively.
We employ a StyleGAN-like architecture to make the intermediate latent space be equipped with rich representational power.
We also introduce a loss function to regularize the intermediate latent codes of two modalities, enabling them to learn a same feature distribution.
Through this strategy, we learn a homogeneous intermediate feature space, $\mathit{W}$, that shares common information of the two modalities, thus producing best results for heterogeneous face recognition. To enforce latent codes in $\mathit{W}$ separable in the angular space, we learn AdaCos~\cite{zhang2019adacos} for the photo-sketch recognition task.

$\mathbf F_{p}$ and $\mathbf F_{s}$ employ a simple encoder architecture that contains convolution, max pooling and fully connected layers. The style generators, $\mathbf G_{p}$ and $\mathbf G_{s}$, consist of several style blocks and deconvolution layers as in~\cite{karras2019style}. However, unlike~\cite{karras2019style}, we do not use noise inputs and progressively growing architecture because the sole purpose of our style generators is to help the homogeneous intermediate latent space retain common representational information of the two modalities for reducing the modality gap between them. 
Our style generator architecture is very light as compared to that of StyleGAN due to limited number of training samples.
The discriminators, $\mathbf D_{p}$ and $\mathbf D_{s}$, distinguish generated photo/sketch and real samples by taking corresponding concatenated photo and sketch. We use PatchGAN architecture~\cite{isola2017image} of 70x70. Unlike the discriminator in~\cite{zhu2019gan}, our discriminator uses \textit{Instance normalization} instead of \textit{Batch normalization}.

\subsection{Loss functions}
The joint loss function used to train our framework is defined as:
\begin{align}\label{equ:1}
 \begin{split}
\mathit{ L = L_{AdaCos} + \lambda_{GAN}L_{GAN} + \lambda_{s}L_{s} + \lambda_{w}L_{w} }
  \end{split}
\end{align}
GAN loss function, $\mathit L_{GAN}$ ~\cite{goodfellow2014generative}, along with the similarity loss, $\mathit L_{s}$, are used to train the bidirectional photo/sketch synthesis part of the whole network. $\mathit L_{GAN}$ helps generating real and natural-looking synthetic outputs while the similarity loss, 
$\mathit L_{s}$, measures pixel-wise $\mathit \ell_{1}$ distance between generated and real photo/sketch images. To regularize and enforce the same distribution for photo, $\mathit w_{p}$, and sketch, $\mathit w_{s}$, in the intermediate latent space, we introduce a collaborative loss, $\mathit L_{w}$. It minimizes $\mathit \ell_{1}$ distance between $\mathit w_{p}$ and $\mathit w_{s}$ of the same identity. We use AdaCos loss function~\cite{zhang2019adacos}, $\mathit L_{AdaCos}$, to learn identity recognition. It measures the angular distance in the $\mathit{W}$ space. It is minimized for intra-class features and maximized for inter-class features.

$\mathit\lambda_{GAN}$, $\mathit\lambda_{s}$, and $\mathit\lambda_{w}$ in Eq.~\eqref{equ:1} control the relative importance of each loss function in the bidirectional photo/sketch synthesis task. We used $\mathit\lambda_{GAN}$ = 1, $\mathit\lambda_{s}$ = 10, and $\mathit\lambda_{w}$ = 1 in our experiments.


\subsection{Training}\label{sec:training}
To overcome the problem of insufficient amount of paired photo/sketch training data, we introduce a simple and effective three-step training scheme as shown in Figure \ref{fig:training}. 
In step 1, we train the bidirectional photo/sketch synthesis network using paired photo-sketch training samples to set up an initial homogeneous intermediate latent space, $\mathit{W}$. We use our joint loss function in Eq.~\eqref{equ:1}, excluding the AdaCos loss function, $\mathit L_{AdaCos}$. 
In step 2, we pre-train the photo mapping network, $\mathbf F_{p}$, using AdaCos loss only on the publicly available large photo database CelebA~\cite{liu2015faceattributes} to overcome the problem of insufficient sketch training samples. 
Then, we train our full network in step 3 using the whole joint loss function in Eq.~\eqref{equ:1} on target photo/sketch samples.

\section{Experiments}
\label{Section4}
\label{exp:collaborative loss}
\begin{figure*}
\begin{center}
\includegraphics[width=0.95\textwidth]{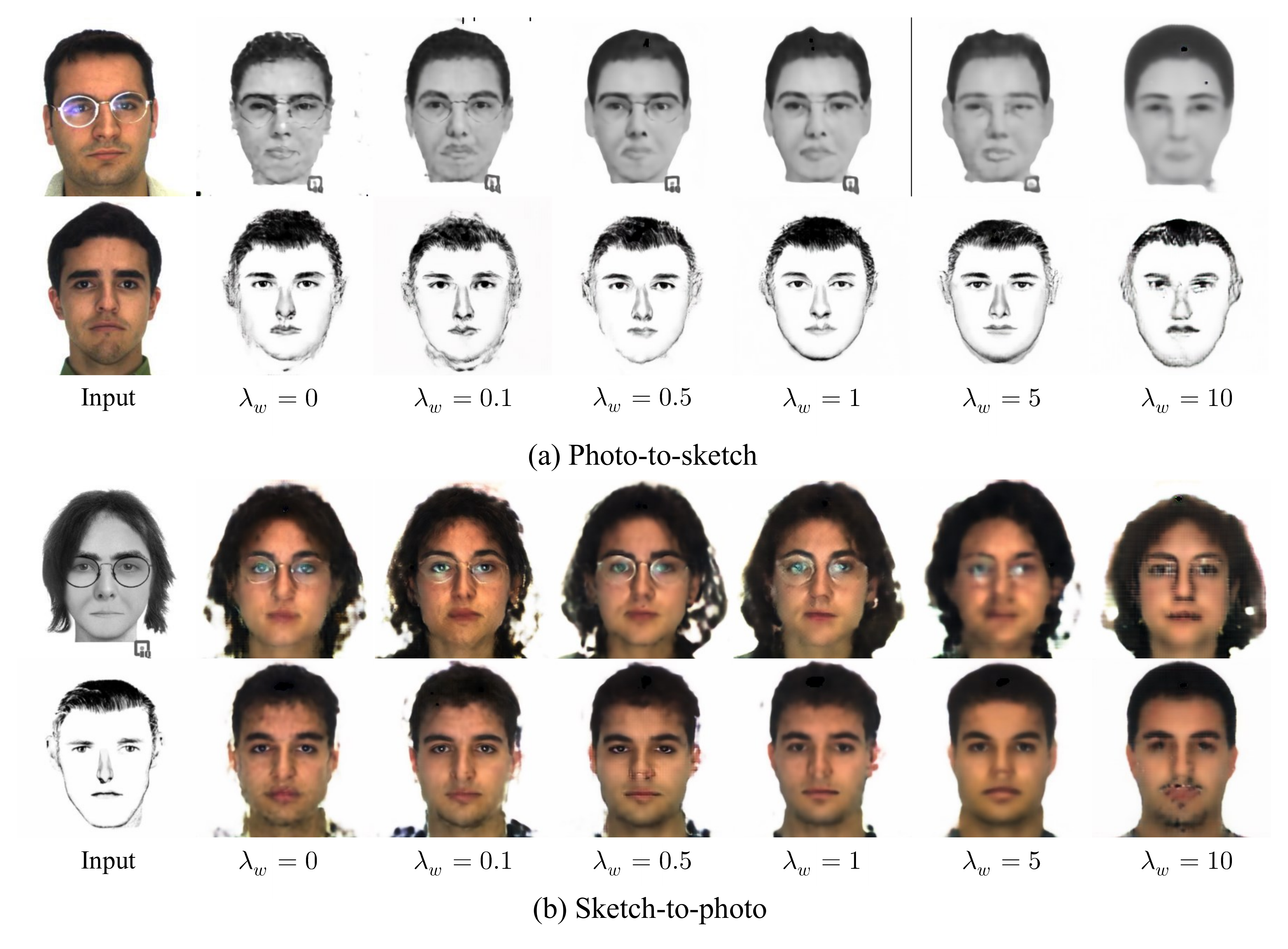}
\end{center}
\vspace{-1.5 em}
   \caption{Synthesis results of our style generators for different values of $\mathit\lambda_{w}$. (a) photo-to-sketch synthesis and (b) sketch-to-photo synthesis. First and second rows are for Faces (In), while third row is for Identikit (As). Images collapse with high $\mathit\lambda_{w}$ so that network could not learn representational information of photo and sketch. $\mathit\lambda_{w}$=1 shows the best synthesis results. (Please view in color.)}
\label{fig:collaborative loss}
\vspace{-0.5em}
\end{figure*}
\subsection{Data description and implementation details}
We have conducted our experiments using the e-PRIP composite sketch database.
The e-PRIP~\cite{mittal2014recognizing} database consists of four different composite sketch sets of 123 identities. However, only two of them are publicly available: the composite sketches created by an Indian user adopting the FACES tool \cite{facetool}, and an Asian artist using the Identi-Kit tool \cite{idtool}. We have used 48 identities for training and the remaining 75 identities for test.

All images are aligned by eye position and initially cropped to 272x272. Then, they are randomly cropped to 256x256 during training. We optimize our network using Adam optimizer with the learning rate of 0.0002 and batch size 8, in step 1 and 3 of training. We use the learning rate 0.0005 and batch size 32 in step 2.
We train our network for 3,000 epochs on the CUFS~\cite{tang2003face} viewed sketch database in step 1 of training, 50 epochs on CelebA~\cite{liu2015faceattributes} in step 2, and 3,000 epochs on the target database in step 3.

The recognition accuracies of our network presented in the following sections are average results over five experiments with random partitions.

\begin{table}[]
\caption{Rank 50 recognition accuracy (\%) on the e-PRIP database with a gallery size 1,500.}
\begin{center}
\begin{tabular}{ccc}
\hline
Method                   & Faces (In)      & Identikit (As)      \\ \hline
Kazemi \textit{et al.}~\cite{kazemi2018attribute}            & 77.50               & 81.50               \\ 
Iranmanesh \textit{et al.}~\cite{iranmanesh2018deep}          & 80.00                  & 83.00                       \\ 
\textbf{Ours}                    & \textbf {93.86}                  & \textbf {90.40}                      \\ \hline
\end{tabular}
\end{center}
\label{tab:eprip_1.5k}
\vspace{-1.5em}
\end{table}

\begin{table}[]
\caption{Rank 50 recognition accuracy (\%) on the e-PRIP database with a gallery size 10,075.}
\begin{center}
\begin{tabular}{ccc}
\hline
Method                   & Faces (In)      & Identikit (As)      \\ \hline
G-HFR~\cite{peng2016graphical}          & -              & 51.22                 \\ 
DLFace~\cite{peng2019dlface}           & 70.00              & 58.93                  \\ 
CAGTL~\cite{liu2020coupled}          & 78.13              & 67.20                  \\ 
\textbf{Ours}          & \textbf {92.78}                  & \textbf {88.26}                 \\ \hline
\end{tabular}
\end{center}
\label{tab:eprip_10k}
\vspace{-2.5em}
\end{table}
\begin{table*}[]
\caption{Rank 50 recognition accuracy (\%) on the e-PRIP database with a gallery size 1,500 for the synthesis network.}
\begin{center}
\begin{tabular}{ccc}
\hline
Method                   & Faces (In)      & Identikit (As)      \\ \hline
Only mapping networks           & 19.74               & 43.72               \\ 
\makecell{Photo-to-sketch (with $\mathbf G_p$ removed)}         & 68.54                & 61.58               \\ 
\makecell{Sketch-to-photo  (with $\mathbf G_s$ removed)}          & 73.84               & 73.88               \\ 
\makecell{Our full network (with both $\mathbf G_p$ and $\mathbf G_s$)}             & \textbf {93.86}                  & \textbf {90.40}                      \\ \hline
\end{tabular}
\end{center}
\label{tab:synthesis network}
\vspace{-1.5em}
\end{table*}

\subsection{Photo-sketch recognition results}
\label{exp:comparison}

In this section, we compare the performance of our method with that of representative state-of-the-art photo-sketch matching methods on the two subsets of e-PRIP dataset~\cite{mittal2014recognizing}.
Let us denote them FACES (In) and Identikit (As), respectively. We perform the experiments with an extended gallery to a mimic real law-enforcement scenario where multiple numbers of suspects are selected from a large photo database. With extended gallery setting, rank 50 accuracy is most commonly used criteria. Thus we compared rank 50 accuracies.
While some photos in extended galleries of previous works are not publicly available, we have tried to mimic their gallery as close as possible using publicly available databases for fair comparison.

Following~\cite{kazemi2018attribute} and~\cite{iranmanesh2018deep}, we have constructed an extended gallery of 1,500 subjects including probe images by using photos from ColorFERET~\cite{phillips2000feret}, Multiple Encounter Dataset (MEDS)~\cite{founds2011nist}, and CUFS~\cite{tang2003face}.
The results are presented in Table \ref{tab:eprip_1.5k} where the accuracies for Kazemi \textit{et al.} and Iranmanesh \textit{et al.} are obtained from their CMC curves. Our method achieved 93.86\% rank 50 accuracy on Faces (In) which was 13.86\% higher than \cite{iranmanesh2018deep}. On Identikit (As), our method achieved 90.40\% which outperformed SOTA.

To compare the performance with~\cite{peng2016graphical, peng2019dlface} and~\cite{liu2020coupled}, we have built another extended gallery of 10,000 subjects using face photos collected from the aforementioned photo databases and the labeled faces in the wild-a (LFW-a) database~\cite{wolf2010effective}. The test gallery set contains the total of 10,075 face photos. Table \ref{tab:eprip_10k} shows the comparison results of our method with the previous state-of-the-art representative methods. As can be seen, our method shows the far better performance of 92.78\% and 88.26\% rank 50 accuracies on Faces (In) and Identikit (As), respectively, with large margins.
These results show that our bidirectional collaborative StyleGAN-like Synthesis Network learns an effective intermediate latent space with rich representational power for face photo-sketch recognition task.  


\subsection{Effect of bidirectional collaborative synthesis of photo-to-sketch and sketch-to-photo}
\label{exp:synthesis network}
To verify the effectiveness of our StyleGAN-like bidirectional collaborative synthesis network on the recognition task, we give comparison with three different versions from the full network.
In the first version, we removed the style generators, $\mathbf G_s$ and $\mathbf G_p$, from the network in Figure \ref{fig:network} and train the mapping networks, $\mathbf F_p$ and $\mathbf F_s$, using AdaCos loss function. That is, the first version could not take any advantage of synthesis network. 
For this version, the mapping networks are pre-trained for 50 epochs on the CelebA photo database~\cite{liu2015faceattributes}, then fine-tuned for 3,000 more epochs on the target database. 
For the second and third versions, we trained a unidirectional synthesis based photo-sketch recognition network by using only one of the style generators, either $\mathbf G_s$ or $\mathbf G_p$. These two versions employed the three-step training scheme as in the full network. 

The comparison results in Table \ref{tab:synthesis network} indicate that the addition of either photo or sketch generator improves the recognition accuracy. 
The unidirectional sketch-to-photo network shows better performance than the unidirectional photo-to-sketch network. 
This is because sketch-to-photo network translates the information-poor input to information-rich output, thus providing better representational feedback to the intermediate latent space as compared to photo-to-sketch network.
However, it still cannot provide enough representational power.
 
Our full network which exploited the bidirectional collaborative synthesis network dramatically improved the recognition performance.
It is because our bidirectional synthesis network warrants the intermediate latent space to have important representational information by utilizing the mutual interaction between the two opposite mappings.

\subsection{Effect of three-step training scheme}
\label{exp:pre-training}

To validate the effectiveness of the proposed three-step training scheme, we compare three different training settings in Table \ref{tab:pre-training}. For this, we train our model 1) using only step 3, that is, without pre-training, 2) using step 2 and step 3, and 3) using all the three steps. 
We can see that there is significant improvement in recognition accuracy when using pre-training (step 2), especially for Faces (In) dataset. 
This shows the power of large-scale pre-training in solving data scarcity problem. The combination of all the three training steps further boosts the recognition performance. Step 1 provides an effective initialization of the intermediate latent space between photo and sketch for large-scale training in step 2.
As the last row in Table \ref{tab:pre-training} shows, our three-step training strategy effectively overcomes the problem of insufficient sketch training samples.


\begin{figure*}
\begin{center}
\includegraphics[width=0.95\textwidth]{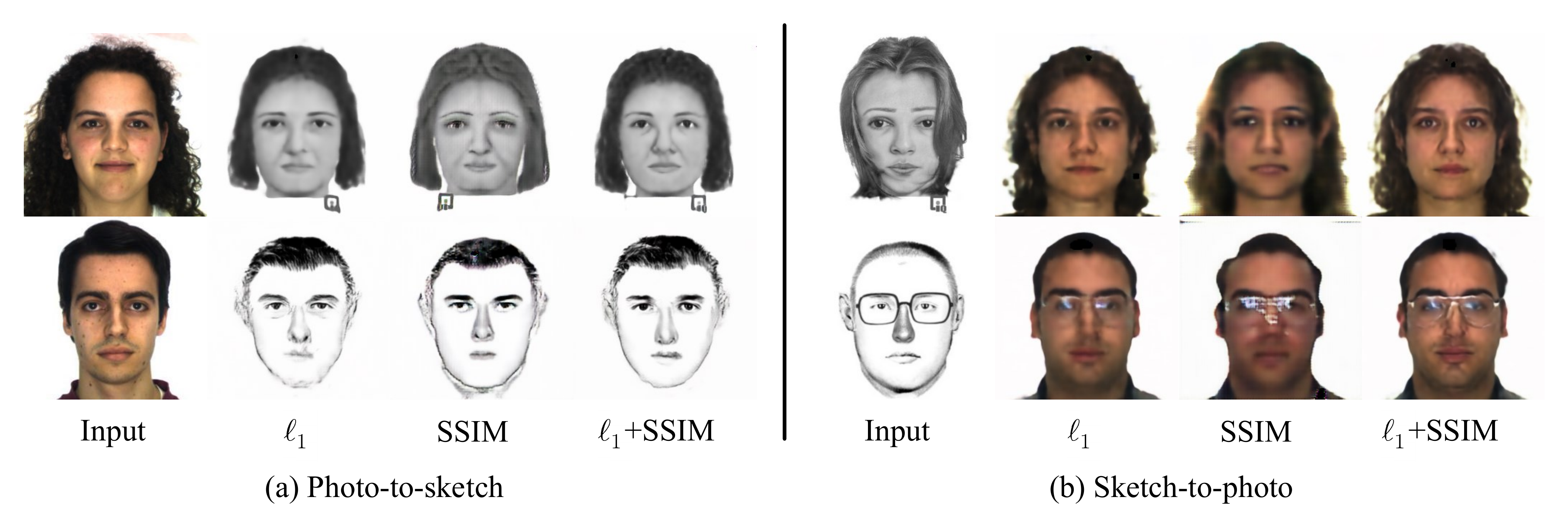}
\end{center}
\vspace{-1.5 em}
   \caption{Synthesis results of our style generators for three different versions of $\mathit L_{s}$. (a) photo-to-sketch synthesis and (b) sketch-to-photo synthesis. First and second rows are trained on Faces (In), while third and fourth rows are trained on Identikit (As). (Please view in color.)
   }
\label{fig:similarity}
\vspace{-1 em}
\end{figure*}
\begin{table*}[]
\caption{Rank 50 recognition accuracy (\%) on the e-PRIP database with a gallery size 1,500 for training scheme.}
\begin{center}
\begin{tabular}{ccc}
\hline
Method                   & Faces (In)      & Identikit (As)      \\
\hline
\makecell{Without pre-training (step 3 only)}    & 25.32   & 46.14   \\ 
\makecell{Two-step training (step 2 + step 3)}     & 90.66   & 89.60 \\
\makecell{Three-step training (step 1 + step 2 + step 3)}     & \textbf {93.86}                  & \textbf {90.40}   \\ 
\hline
\end{tabular}
\end{center}
\label{tab:pre-training}
\vspace{-2.5em}
\end{table*}
\subsection{Collaborative loss, $\mathit L_{w}$}
In this section, we analyze the effect of collaborative loss, $\mathit L_{w}$, on the recognition accuracy.
We experimented our network as we change the value of $\mathit\lambda_{w}$. Table \ref{tab:collaborative loss} shows the results for different values of $\mathit\lambda_{w}$  on the extended gallery setting of 1,500 samples.

The performance for $\mathit\lambda_{w}=0$ is poor. $\mathit\lambda_{w}=0$ means that our network is not using collaborative loss $\mathit L_{w}$. The network is unable to constrain the two mappings symmetrical.
The accuracy improves when we increase the value of $\mathit\lambda_{w}$ as can be seen in Table \ref{tab:collaborative loss}.
Through many experiments, we have found that $\mathit\lambda_{w}$ = 1 produces the best result for our task.
These results show that our collaborative loss helps regularizing the intermediate latent representations of the two different modalities, effectively aligning the two modalities in the intermediate latent space. However, as $\mathit\lambda_{w}$ gets too large, the performance degrades as can be seen in Table \ref{tab:collaborative loss}.
We think that a large $\mathit\lambda_{w}$ emphasizes too much on making latent codes symmetrical, and breaks the learning balance of the latent space between representational capcacity and symmetrical mapping.

Figure \ref{fig:collaborative loss} shows examples of synthesis results produced by our style generators for different values of $\lambda_{w}$. There is a general trend that better synthesis results yield better recognition accuracies.
For $\mathit\lambda_{w}$ = 10, the results collapsed to the same synthesis result for most of the target samples. This shows that too much weightage to the collaborative loss strongly enforces the same latent distribution while the representational capacity of the latent space relatively ignored.

\begin{table}[]
\caption{Rank 50 recognition accuracy (\%) on the e-PRIP database with a gallery size 1,500 for $\lambda_{w}$.}
\begin{center}
\begin{tabular}{ccc}
\hline
Method                   & Faces (In)      & Identikit (As)      \\ \hline
$\mathit\lambda_{w}$ = 0            & 72.00               & 66.40               \\ 
$\mathit\lambda_{w}$ = 0.1           & 89.32                  & 82.68                       \\ 
$\mathit\lambda_{w}$ = 0.5           & 89.60                  & 85.60                       \\ 
$\mathit\lambda_{w}$ = 1                     & \textbf {93.86}                  & \textbf {90.40}                      \\ 
$\mathit\lambda_{w}$ = 5            & 85.34               & 84.28               \\ 
$\mathit\lambda_{w}$ = 10            & 83.72               & 83.74               \\ \hline
\end{tabular}
\end{center}
\label{tab:collaborative loss}
\vspace{-1.5em}
\end{table}

\subsection{Similarity loss, $\mathit L_{s}$}
\label{exp:similarity loss}
Figure \ref{fig:similarity} shows the results produced by our style generators for three simple variations of $\mathit L_{s}$. 
First, we used pixel-wise $\mathit \ell_{1}$ distance only as our $\mathit L_{s}$. Second, we used only patch-wise structural similarity (SSIM) loss \cite{snell2017learning}. Third, we employed SSIM loss along with $\mathit \ell_{1}$ distance for $\mathit L_{s}$. Figure \ref{fig:similarity} shows that using only SSIM loss for $\mathit L_{s}$ produces the worst synthetic results, yielding the lowest recognition accuracy as can be seen in Table \ref{tab:similarity loss}.
On the other hand, $\mathit \ell_{1}$ produces the best recognition results compared to the other two settings.
Our observation is that SSIM loss provides extra structural information for synthesis, but it does not help for recognition task.
Thus, we opt to use only $\mathit \ell_{1}$ distance as our $\mathit L_{s}$ in the joint loss function in Eq.~\eqref{equ:1}.

\begin{table}[]
\caption{Rank 50 recognition accuracy (\%) on the e-PRIP database with a gallery size 1,500 for $\mathit L_{s}$.}
\begin{center}
\begin{tabular}{ccc}
\hline
Method                   & Faces (In)      & Identikit (As)      \\ \hline
$\mathit \ell_{1}$           & \textbf {93.86}                  & \textbf {90.40}               \\ 
SSIM           & 81.86                  & 79.74                       \\ 
$\mathit \ell_{1}$ + SSIM           & 91.98                  & 89.34                       \\ \hline
\end{tabular}
\end{center}
\label{tab:similarity loss}
\vspace{-2.5em}
\end{table}

\section{Conclusion}
We proposed a novel deep learning based face photo-sketch recognition method by exploiting a homogeneous intermediate latent space between photo and sketch modalities. For this, we introduce a bidirectional photo/sketch synthesis network based on a StyleGAN-like architecture.
In addition, we employ a simple three-step training scheme to overcome the problem of insufficient paired training samples. The experiment results have verified the effectiveness of our method, outperforming the representative state-of-the-art methods. Our method shows great promise in matching pairs of other different modalities.


\clearpage
\printbibliography
\end{document}